# Representing and Combining Partially Specified CPTs

Suzanne M. Mahoney
Information Extraction and Transport, Inc.
1911 Fort Myer Drive
Arlington, VA 22209
suzanne@iet.com

Kathryn Blackmond Laskey
George Mason University
4400 University Drive
Fairfax, VA 22032
klaskey@gmu.edu

## Abstract

This paper extends previous work with network fragments and situation-specific network construction. We formally define the asymmetry network, an alternative representation for a conditional probability table. We also present an object-oriented representation for partially specified asymmetry networks. We show that the representation is parsimonious. We define an algebra for the elements of the representation that allows us to 'factor' any CPT and to soundly combine the partially specified asymmetry networks.

## 1. INTRODUCTION

A Bayesian network represents the probabilistic dependencies among a set of random variables by a directed acyclic graph (Pearl, 1988). Each node in the network represents a random variable (a set of mutually exclusive and exhaustive possible values for some hypothesis). A node in a Bayesian network is conditionally independent of all non-descendant nodes given its direct parents. Each node in the network stores its probability distribution given its parents. This information is sufficient to represent implicitly the full joint probability distribution over all nodes in the network. Efficient inference algorithms have been developed for computing the conditional distributions for each node given observed values of some nodes in the network.

The graphical structure of Bayesian networks makes explicit certain dependencies and independencies among the variables in the network. The standard representation for the conditional distribution of a node given its parents is a *conditional probability table* (CPT), which specifies a probability distribution for values of the node for each unique combination of parent states. Frequently, there are other structural relationships among the distributions in the conditional probability tables (CPTs). These relationships are represented only implicitly in the standard representation. A number of authors have suggested structured representations that capture additional semantic information about the conditioning distributions. These include independence of causal influence models such as the noisy-OR (Pearl, 1988; Srinivas, 1993), parametric models such as the sigmoid function (Jaakkola and Jordan, 1996a; Neal, 1992), and models for asymmetric independence such as multinets (Geiger and Heckerman, 1991) or decision trees (Boutilier, et al. 1996). Such structural information, when available, can greatly reduce the burden of knowledge elicitation and can often be exploited to increase the efficiency of inference.

In this paper, we propose a general representation for *asymmetric independence* relationships. Asymmetric independence occurs when a variable is independent of another variable for some configurations of its parent variable(s) but not others. When there is asymmetric independence, the standard CPT representation fails to capture structural knowledge implicitly encoded as equality relationships among distributions in the CPT. One common type of asymmetric independence is *subset independence* (Heckerman, 1990; Geiger and Heckerman, 1991), which occurs when the distribution of the child is independent of one parent given a subset of the values of another parent. The subset for which the distributions are the same may represent possible causes of a common effect. Subset independence relationships can be represented by specifying probability distributions not for each unique combination of parent states, but for each element of a *partition* of the cross product of the parents' state spaces.

A number of belief network software packages now support partitions. In his local expression language for probabilistic dependencies, D'Ambrosio (1995) defines rectangular partitions that are the cross-product of partitions of parent variables. Boutilier et al (1996) defined *context-specific independence* (CSI), in which independence relationships exist in some contexts (combinations of states of parent variables) but not others. They used trees to represent the conditioning structure of a CPT and showed how to exploit context-specific independence for efficient inference. Poole (1998) defined *minimal parent contexts* that expressed the conditioning for a given distribution by a minimum set of parent values. Shimony (1993) defined disjunctive assignments for nodes when the probability distributions of the descendent nodes were statistically independent of which value in the disjunction was conditioned on. Friedman and Goldszmidt (1996) use a decision tree to represent local structure in a conditional probability table to facilitate and improve learning of the Bayesian networks.

In this paper we will define asymmetry networks, an alternative representation for CPTs that is more expressive than the tree representation or the rectangular partition. Asymmetry networks can be used to represent subset independence, context-specific independence, or any independence relationship involving equality constraints among distributions in the CPT. Asymmetry networks are a natural way to specify the partial conditional distributions for network fragments (Laskey and Mahoney, 1997). Network fragments are an object-oriented representation designed to support construction of situation-specific networks (Mahoney and Laskey, 1998). A network fragment holds a partial representation of the distribution of its resident variables. Network fragments factor an asymmetry network in a way that provides an explicit representation for subset independence relationships and reveals context-specific independence. In this paper we define asymmetry networks and demonstrate their utility as a parsimonious representation for asymmetric independence. We describe how context-specific independence can be exploited to decompose an asymmetry network into subnetworks associated with hypothesis-conditioned network fragments. We describe operators for consistently combining the asymmetry subnetworks associated with hypothesis-specific network fragments during construction of a situation-specific network.

## 2. BACKGROUND

### 2.1 NETWORK FRAGMENTS

*Network fragments* (Laskey and Mahoney, 1997) provide an object-oriented representation for probabilistic knowledge. A probability model is represented implicitly as a knowledge base of Bayesian network fragments. Each fragment consists of a set of random variables connected by a *fragment graph*, together with information used to construct local distributions for variables. Variables in a fragment are classified as resident, input, or hypothesis variables. *Hypothesis variables* represent the context in which the independence relationships in the fragment hold. *Input variables* condition the distributions of the resident variables, but have their own distributions defined in other fragments. The fragment represents information needed to construct the distributions of its *resident variables*. Context-specific independence relationships can be specified by identifying a set of hypothesis variables, partitioning their state spaces into a set of contexts in which independence relationships hold, and defining a *hypothesis-conditioned fragment* for the remaining variables given each context.

Both fragments and random variables are objects organized in a type hierarchy, with associated structure and methods. Each random variable and fragment has a set of *identifying attributes* which, when specified, create a unique identifier for an instance of the random variable or fragment. There is a mapping from identifying attributes of the fragment to identifying attributes of its random variables. Identifying attributes play the role of variables in a logic programming language (to avoid confusion, we reserve the term variable for random variables).

When a variable's probability distribution has local structure, it is often convenient to specify its distribution in several different fragments to be combined at run time into a full distribution. For example, the "Disease" node in a medical diagnosis system may be specified as a noisy-OR distribution with perhaps hundreds of input diseases. These input distributions could be organized into groups of related diseases, each represented as a separate fragment. At run-time there may be information available that rules out entire categories of diseases, requiring only a small percentage of the groups to be included in the final constructed model.

We use influence combination to represent local structure (Laskey and Mahoney, 1997). Random variables have an *influence type* with associated influence combination method. Enabling conditions for a given influence type specify restrictions on the influencing variables. For example, a noisy-OR variable must take only binary input variables. Each individual fragment uses an *influence function* to represent parameters of the *influence combination method*. For example, the trigger probability for a candidate cause in a noisy-OR distribution is returned by the influence function in the fragment where the cause-effect link is defined. In this paper we focus on local structure arising from asymmetric independence relationships.

### 2.2 NETWORK CONSTRUCTION AND FRAGMENT COMBINATION

In complex domains it is infeasible to construct a complete Bayesian network encompassing all the situations one might encounter in problem solving. Knowledge-Based Model Construction (KBMC) is the process of constructing a model for a problem instance from a knowledge base representing generic domain entities and their interrelationships. A KBMC system includes a knowledge base, search operators for retrieving problem-relevant knowledge base elements, network construction operators, network evaluation operators, and model construction control mechanisms. Objectives for a KBMC system are to minimize costs of representation, retrieval, construction and evaluation, while providing accurate responses to queries.

In previous work (Laskey and Mahoney, 1998), we proposed a model construction control strategy for producing situation-specific networks from a knowledge base of network fragments. A situation-specific network is a minimal network sufficient to respond to a query for which the knowledge base is query complete. During network construction, creation of random variable instances triggers the retrieval of network fragments in which the corresponding random variables are resident. The identifying attributes of the random variable in a retrieved fragment are unified with the values of the identifying attributes of the instance that triggered its retrieval. These identifying attribute values are also unified with the same identifying attributes appearing in other random variables in the fragment. This process creates new random variable instances, which triggers the retrieval of fragments in which they are resident.

After a retrieved fragment is unified with its identifying attributes, variables in the retrieved fragment are combined with matching variables in the situation-specific network. When fragments are combined, local distributions for the combined fragment are computed from the fragment influence functions using the node's influence combination method. The influence function for a variable in a fragment in which it is resident must provide the inputs needed by that variable's influence combination method. Thus, influence combination and influence functions must be designed to work together. We cataloged (Laskey and Mahoney, 1997) a number of generic influence combination methods: Simple-Combination, independence of causal influence (ICI), and Parameterized-Combination. We also defined the requirements for a combination method that combines hypothesis-conditioned fragments representing asymmetric independence relationships.

## 3. ASYMMETRY NETWORKS

### 3.1 DEFINITION

An asymmetry network represents asymmetric independence relationships for a conditional probability table in a Bayesian network. An asymmetry network has three components: 1) a random variable for which a distribution is being defined; 2) a partition of the state space defined by the parents of the random variable; and 3) a function from the partition elements of the parents' state space to distributions for the dependent variable. Whereas the tree representation is limited to rectangular partitions, an asymmetry network can represent an arbitrary partition of the state space defined by the parent variables.

**Definition 1** Let $\Pi$ be a set of random variables with state space $\sigma_\Pi$. Let $R$ be a partition of $\sigma_\Pi$. Let $X$ be a random variable such that $X \notin \Pi$. Let $H$ be a set of distributions for $X$. Let $F$ be a function from $R$ to $H$. Let $\rho_j$ be a partition element of $R$. If $F(\rho_j)$ is defined for all partition elements $\rho_j \in R$, then $F$ is known as a *asymmetry network function* for $X$ and $R$. If $\pi$ is empty and $F$ is defined for the null set, then $F$ is simply an asymmetry network function for $X$.

The partition $R$ defines a partitioning of the conditioning events for the random variable into subsets for which the distribution is the same. The asymmetry network function maps each partition element into a probability distribution for $X$ that applies for all parent variable configurations in the partition element. In our experience, the same partition R can be used as the domain of the asymmetry function for several different child variables. Because $F$ maps a partition element to any positive number, an asymmetry network function may be defined in terms of ratios and frequencies, which convey semantic meaning, as well as probabilities. In our network fragment representation, we have specified an influence type called asymmetry_variable, for which the influence function is an asymmetry function.

**Definition 2** Let $\Pi$ be a set of random variables. Let $R$ be a partition of the state space $\sigma_\Pi$. Let $X$ be a random variable such that $X \notin \Pi$. Let $F$ be an asymmetry network function for $X$ and $R$. Then $N = (R, X, F)$ is an *asymmetry network* for $X$, with conditioning partition R and asymmetry function F.

An asymmetry network represents a conditional probability distribution for a dependent variable. It is often convenient to decompose an asymmetry network into a set of *asymmetry subnetworks*. An asymmetry subnetwork maps a subset of the associated asymmetry network's conditioning partition onto distributions for the dependent variable. The distributions of the asymmetry subnetwork must be consistent with those of the associated asymmetry network.

**Definition 3** Let $N = (R, X, F)$ be an asymmetry network. Let $\rho_k$ be a partition element of $R$. Then $N_j = (R_j, X, F_j)$ is an *asymmetry subnetwork* of N if $R_j \subseteq R$, $R_j$ is the domain of $F_j$, and $F_j(\rho_k) = F(\rho_k)$ for all $\rho_k \in R_j$. $R_i$ is called the conditioning partition for the subnetwork and $F_i$ is called the asymmetry function.

### 3.2 ASYMMETRY NETWORKS EXAMPLE

Consider the following example that illustrates how an asymmetry network encodes equality constraints among the conditional probability distributions of a variable. The task is to model the activity of military unit given a set of environmental conditions. Background information is as follows: (1) The unit does not want to be detected by airborne photographers while it is moving. (2) Nor does it wish to be on the road when driving may be hazardous. Figure 1 presents a graphical representation of the asymmetry network for the "Activity" of the unit. The shaded boxes represent the partition elements of an asymmetry partition for "Activity" and its parents, "Wind", "Rain" and "Time of Day". Each partition element is a set of elements from the state space of the parent variables. The white boxes to the right are the associated conditional probability distributions. A statement summarizing the rationale used to determine the elements that appear in the partition element is below each partition element-distribution pair. The key for the variables' states is at the lower right hand side of the picture.

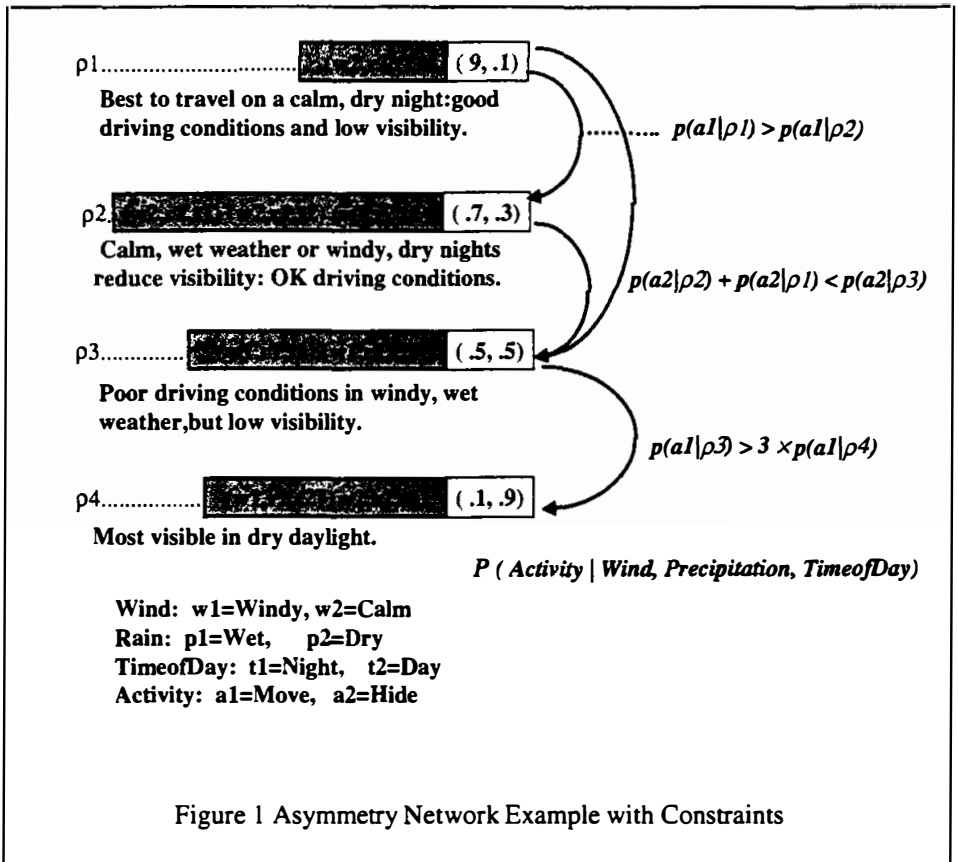

Figure 1 Asymmetry Network Example with Constraints

In addition to the equality constraints encoded in the asymmetry network, there often exist semantic relationships between distributions for different partitions. In Figure 1, annotated arrows link the partition element - distribution pairs. These annotations represent constraints that hold among the linked partition element-distribution pairs. The annotated arrow between the top two partition element-distribution pairs represents the knowledge: the probability of the unit moving during a calm dry night is greater than the probability of the unit moving during wet weather or windy, dry nights. The annotation on the merged arrows from the top two pairs to the third pair represents the knowledge: the probability of hiding during windy wet weather is greater than the sum of the probability of hiding during a calm, dry night and the probability of hiding during calm wet weather or windy dry night.

As the example illustrates, asymmetry networks with their paired conditional probability distributions represent structural knowledge about the domain. The partition elements explicitly represent equality constraints among the distributions of the dependent variable. The numbers in the distributions reflect other constraints among the distributions, which can be captured as annotations to the distributions. While constraints other than equality ones are difficult to represent graphically, mathematical expressions represent some of these relationships rather nicely. The example illustrates only a few of the relationships that may exist among distributions. Druzdzel and van der Gaag (1995) catalog a set of constraints that apply to the probabilities of a joint probability distribution. Wellman (1990) catalogs relationships among variables in a qualitative probabilistic network. We hypothesize that many of the constraints tallied in these catalogs can be mapped to constraints among distributions in an asymmetry network. Furthermore, we claim that partition element-distribution pairs are an ideal foundation for a representation that expresses these constraints.

### 3.3 REPRESENTING ASYMMETRY SUBNETWORKS IN NETWORK FRAGMENTS

We use asymmetry subnetworks to represent partial influence relationships among variables in a network fragment. We have the following objectives for the decomposition of an asymmetry network into asymmetry subnetworks: 1) minimize the amount of information that must be represented in a subnetwork; 2) support combination of asymmetry subnetworks; 3) maximize reuse of the representation. Asymmetry networks are associated with network fragments in which their associated random variables are resident.

Both asymmetry networks and asymmetry subnetworks have three parts: a partition of the state space of a set of conditioning variables, a dependent variable and a function from the partition elements of the partition to probability distributions over the states of the dependent variable. The objective is to define an asymmetry subnetwork so that independence relationships hold within the subnetwork that may not hold globally. For example, a variable Y may be a parent of the variable X, but may be independent of X and all other variables in a fragment, given the context of the fragment. In such a case, Y need not be represented in the fragment.

### 3.4 AN EXAMPLE

Figure 2 graphically illustrates how our representation can be used to represent the structural information associated with equality constraints among distributions in a CPT. It shows both the elements of the conditioning structure and the object classes that represent them.

The figure shows two elements of an asymmetry network for the dependent variable W. W has parents X, Y and Z. In this example, parent variable Z has two contexts, $Z=z_1$ and $Z=z_2$. (More generally, these contexts could be subsets of the state space of Z rather than single values.) An asymmetry subnetwork is defined for each of these contexts. The first subnetwork has the context variable Z set to $z_1$. It has a single conditioning partition, marked XY-1. The conditioning partition for this subnetwork consists of two partition elements, represented graphically by the white and gray regions in the rectangle below the designation $Z=z_1$. The first of these partition elements is represented in the figure, and is labeled XY-1.1. The elements of the partition element are $(x_1, y_1)$, $(x_2, y_4)$, and $(x_3, y_4)$. Because the context $Z=z_1$ is understood, we do not explicitly represent it in the conditioning partition. In the second asymmetry subnetwork, Z is set to $z_2$. Its partition is marked XY-2. Again, the asymmetry subnetwork consists of two partition elements, the first of which is represented in the figure and is marked XY-2.1. Again there is a graphical representation of the represented elements: $(x_1, y_2)$, $(x_1, y_4)$, $(x_2, y_1)$, and $(x_3, y_1)$.

In the illustration, the names of the classes being represented are in the right hand column. The symbol ⊗ stands for cross-product. The symbol ⊕ indicates collect. At the bottom of the figure we have a row of simple assignments. Simple assignments are the atomic values for a variable and are identified by unique names. Note that each of these is a singleton element. The next higher row of rectangles represents simple partition elements. A simple partition element is collection of possible values for a single variable. An example of such a collection is [x2 ⊕ x3]. Conditioning assignments, the ellipses in the third row from the bottom, represent the cross-product of partition elements. The conditioning assignment on the left, XY-1, represents (x1, y1). The next conditioning assignment, XY-2, represents (x2, y4) and (x3, y4). These are simply the elements of the cross product of the simple partition elements for the variables X and Y.

Conditioning partitions and their conditioning partition elements represent the structure of partitions of the state space for a set of conditioning variables. A conditioning partition may have multiple conditioning partition elements. Associated with each element is a distribution for the dependent variable. The black boxes in Figure 2 make this association and so represent the asymmetry network function for the dependent variable W. We call the function an *asymmetry mapping*. The key advantage of using an asymmetry mapping to represent distributions is that it indirectly associates a conditioning partition element with a distribution. Neither the conditioning partition element nor the distribution needs to know about the relationship. This facilitates reuse of the conditioning partition elements. Similarly, we associate the asymmetry network with a particular dependent variable's local distribution by simply having the local distribution point to a set of asymmetry mappings. These indirect relationships with the dependent variable and among an asymmetry network's components facilitates reuseability and maintenance of the knowledge base. For instance, if a second dependent variable uses the same conditioning partition or even just some of the same conditioning partition elements, the conditioning partition elements can be readily reused by simply having a different set of asymmetry mappings point to them.

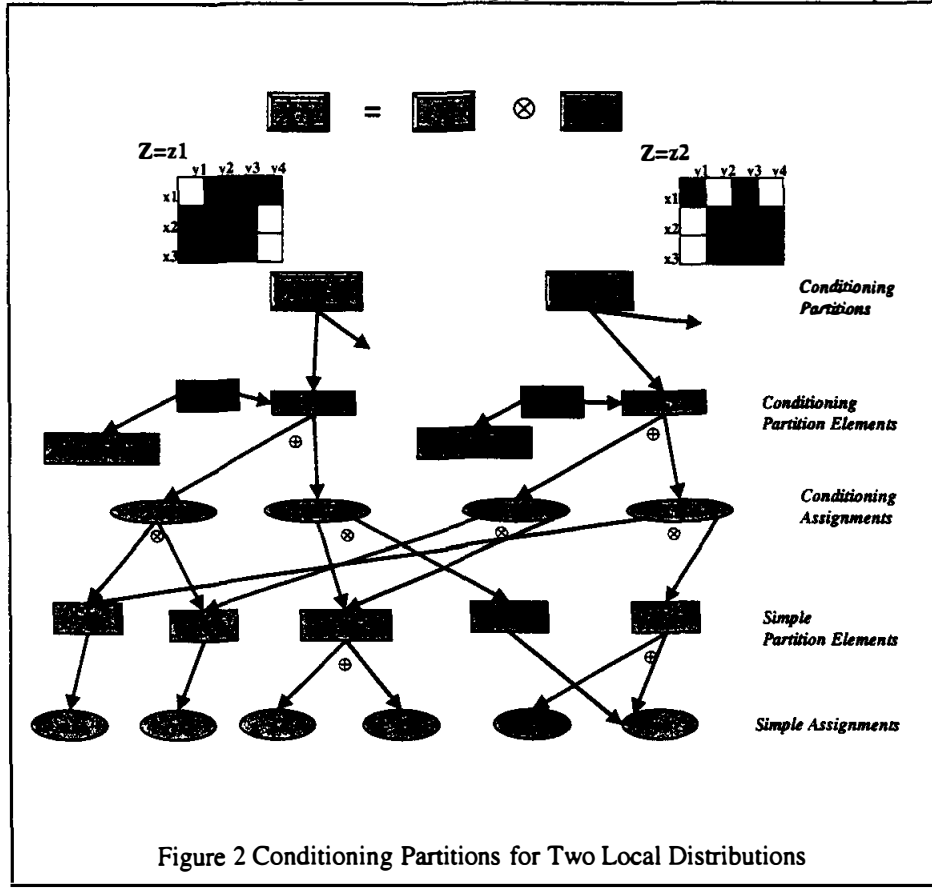

Figure 2 Conditioning Partitions for Two Local Distributions

Consider Figure 2 again. Note that a given simple assignment may contribute to many simple partition elements and that a simple partition element may be part of many conditioning assignments. In turn each conditioning assignment may be referenced by many conditioning partition elements and conditioning partition elements may participate in many conditioning partitions. In addition, our experience is that several different local distributions will make use of the same conditioning partition. For this reason, it makes sense to have an explicit representation for partitions, partition elements and assignments. When creating a local distribution for a resident variable, we can simplify knowledge base construction by reusing or modifying existing structures. When deleting a distribution, we may not automatically delete the associated partitions if they have semantic meaning and may be reused.

In Figure 2, we illustrated portions of two asymmetry subnetworks. One was for the case $Z=z_1$ and the other for $Z=z_2$. In neither case do we explicitly represent the value of Z in the partition. Instead, the value of Z is held fixed as a context value. When constructing a situation-specific network, it may be the case that both values of the context variable Z are relevant. Then, we combine the asymmetry subnetworks, making the context variable explicit. Figure 3 illustrates a portion of the combined asymmetry network. In the combined asymmetry network we retain elements of the old asymmetry networks. To bring the context for Z into the local distribution, we simply construct a conditioning partition element for all three variables. For example, the ellipse XYZ-1 represents a conditioning assignment for the cross product of the conditioning partition element XY-1.1 and the context value $z_1$. The rectangle XYZ-1 is the corresponding conditioning partition element. Note that the same distribution, *Distribution 1*, is now associated through an asymmetry mapping with the rectangle XYZ-1 instead of XY-1.1.

This representation for asymmetry networks has the following advantages. First of all it only represents what actually needs to be explicit for the asymmetry network. In the examples of Figure 2, the context value for Z does not have to be explicitly represented in the asymmetry network because it is pegged to either $z_1$ or $z_2$ for all of the assignments being represented. Only when we combine the representations is the value of Z represented explicitly in the conditioning partitions. The combined representation for asymmetry networks is parsimonious in that it represents only what is necessary to represent, and facilitates reuse of the partition elements of the contributing asymmetry networks. Note that both the new asymmetry mappings and conditional partition elements simply point to elements of the contributing asymmetry networks. Therefore, constructing the combined asymmetry network effectively reuses representation from the contributing asymmetry networks. This permits the efficient maintenance of both the combined and uncombined asymmetry networks in a common knowledge base.

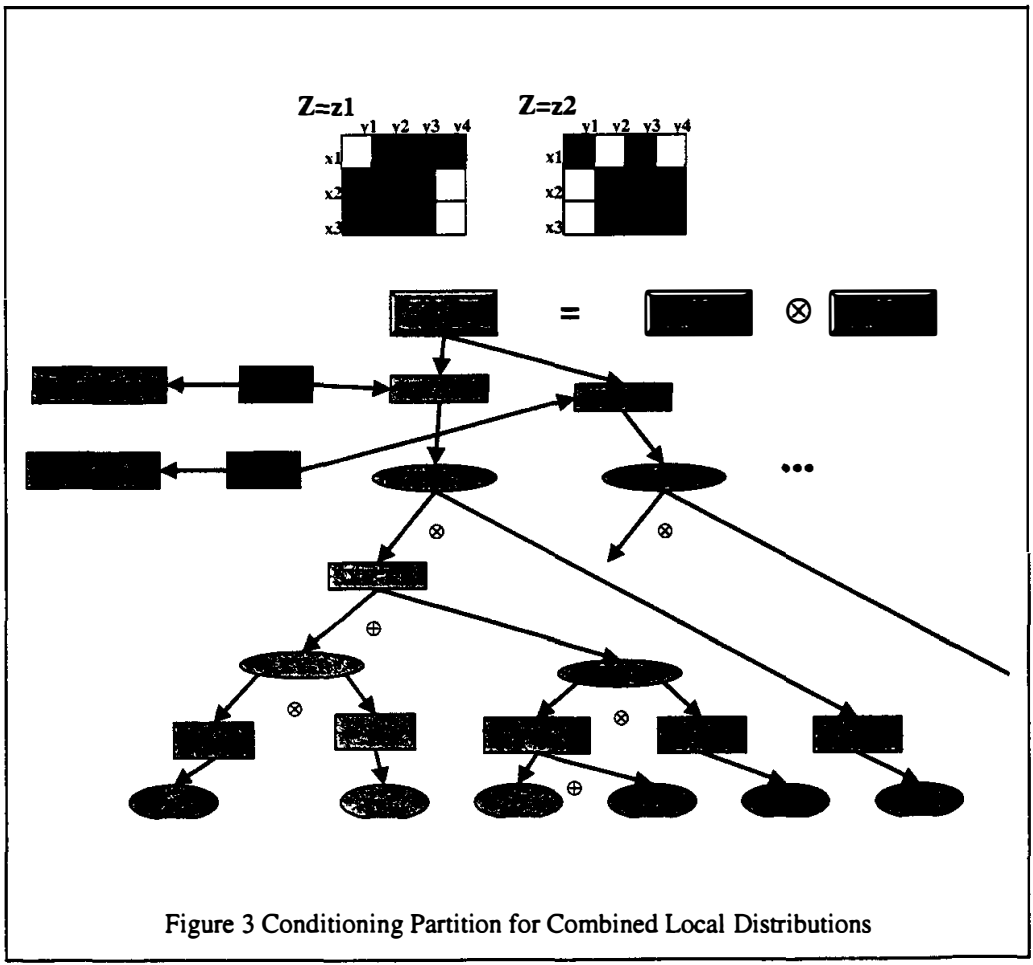

Figure 3 Conditioning Partition for Combined Local Distributions

Context-specific

independence is represented rather nicely in a conditioning partition. In our above example, the dependent variable only depends upon conditioning values of X, Y and Z. These three variables may not be the only parents of the dependent variable W in the global network. The variable W may have another parent V which is applicable in some other context, say Z∈{$z_3$, $z_4$}. In this case, the variable V need not be represented in the network fragments for contexts $z_1$ and $z_2$, but is represented in the fragment for the context {$z_3$, $z_4$}. When this context is combined with contexts $z_1$ and $z_2$, the variable V would appear as a parent to W. It is implicitly represented in the $z_1$ and $z_2$ fragments by its *whole partition element*, a partition element containing all of the possible values for V. In a tree representation (Boutilier, et al, 1996; Friedman and Goldzsmidt, 1996) a branch is simply the cross-product of its variables' values. Simple partition elements represent partitions of single variables (Geiger and Heckerman, 1991; D'Ambrosio, 1995). D'Ambrosio's rectangular partitions are cross-products of Simple Partitions.

## 4. CONTEXT ALGEBRA

This section describes an algebra for 'factoring' and combining the elements of partial conditional probability tables. It is called *context algebra* because it facilitates reasoning about the context elements in network fragments. Like any algebra it has elements and operators. Context algebra's elements represent the classes introduced in Figure 2. Its operators are collect, ⊕, cross, ⊗, and Amap. Cross is used to form cross-products and collect to aggregate elements of the same type. The Amap operator is the 'black box' that associates distributions with conditioning partition elements.

First, context algebra represents both asymmetry subnetworks and the local distributions in hypothesis specific network fragments. Second, context algebra provides a sound method for combining asymmetry subnetworks and the local distributions in hypothesis specific network fragments.

### 4.1 VARIABLE AND DISTRIBUTION TYPES

Before we define the elements and operators of the algebra we define variable types and distribution types. The types are vital to the sound functioning of the algebra. We not only have to distinguish random variable X from random variable Y, we also have to be able to distinguish among instances of random variables. For example, we may want to reason about the Activity of entity E at time t2 given the Activity of entity E at time t1. These would be represented as two different instances of the Activity variable with different values for the identifying attribute time.

We uniquely identify fragment variables by a combination of their random variable and the variable-value pairs for their identifying attributes. We use a variable-value pair in which the value is null to represent an identifying attribute whose value is unspecified. A *variable-value* is the concatenation the name of a random variable with the name of a state from the state space of the specified variable. All instances of a specified random variable that have the same set of variable-value pairs for their identifying attributes are of the same type. A *variable-type* is the name of a random variable concatenated with a set of variable values for each of its identifying attributes. A typed variable is usually denoted by a capital letter: for example X, Y, Z. If two typed variables have the same random variable and different values for their identifying attributes, they carry distinguishing subscripts in parentheses: for example if fragment variable X is instantiated for times t1 and t2, we write $X_{(t1)}$ $X_{(t2)}$. We represent the possible values of a typed variable by subscripted small letters: for example $x_1$, $y_k$, or $z_{(t1)k}$ where k denotes the $k^{th}$ possible value of the random variable.

We also type the distributions. A *distribution-type* is the variable type of the resident variable with which it is associated. We denote a distribution with a bold italic *D* whose first subscript specifies the typed variable. For example, $D_{Xj}$ denotes the $j^{th}$ distribution for the random variable X. If necessary, we may include identifying attributes in parentheses following the variable name: e.g. $D_{X(t1)j}$.

### 4.2 ELEMENTS AND OPERATORS

We begin with a general definition for the elements and operators of our algebra.

**Definition 4** An *element* is an object. It has one or more *operators* that take elements as parameters to construct new elements. An element has an *element-type*. *Simple elements* cannot be subdivided into simpler elements. *Complex elements* are composed of other elements.

The term brace represents subsets of a state space. In a Bayesian network, the parents of a dependent variable form a state space. A brace is associated with a distribution for the dependent variable. A set of braces and their associated distributions represents the local distribution for the dependent variable. *Simple braces* represent *Simple_Assignments*. A *Simple_Assignment* is the atomic representation for the states of a variable.

**Definition 5** A *simple brace* is a simple element. It contains a variable-value pair and an element-type. Its element-type is the variable-type for the variable in the variable-value pair.

Usually we denote a simple brace with a subscripted lower case character. For example, $y_k$ is the $k^{th}$ partition element in a state space for fragment variable Y. If the identifying attributes, $\psi$, of Y need to be made explicit, then the notation is $y_{(\psi)k}$. We also have simple elements for carrying distributions and their types.

**Definition 6** A *typed distribution* is a simple element. It contains a *Distribution* and a distribution-type.

Combination operators construct complex elements from pairs of elements. Map operators pair elements with typed distributions.

**Definition 7** A *combination operator* constructs a complex element from a pair of elements. The combination operator also establishes the element-type for the complex element.

A complex brace is simply a composition of braces. Obviously, a complex brace can be recursively decomposed to reveal a set of simple braces.

**Definition 8** A *complex brace* is a complex element obtained by combining a pair of braces with a combination operator. A type method returns the element-type for the brace.

Consider the example shown in Figure 2. The *Simple_Partition_Element* $[x_2 \oplus x_3]$ is a complex brace. Its element-type is the same as that for the variable X. The complex brace, $([x_2 \oplus x_3] \otimes y_4)$, consists of a complex brace, $[x_2 \oplus x_3]$, and a simple brace, $y_4$, joined by the combination operator $\otimes$. Its element-type is a set of types, one for variable X and the second for variable Y. *Conditioning_Assignments* are usually braces like this one. Obviously, a complex brace can be recursively decomposed to reveal a set of simple braces.

Mappings are elements that contain distributions. Map operators join other elements with distributions. Note that the map operator works only with a single brace and a single distribution.

**Definition 9** A *map operator* constructs an asymmetry mapping from a brace and a typed distribution. It also establishes the element-type for the asymmetry mapping.

**Definition 10** An *asymmetry mapping* is a complex element. It contains a brace, a typed distribution, their map operator, and an element-type. The element-type is a pair consisting of the element-type for the brace and distribution-type for the typed distribution.

See Figure 2. The 'black boxes' represent asymmetry mappings. For example, the *Conditioning_Partition_Element* XY-1.1 is represented by the brace, $[([x_2 \oplus x_3] \otimes y_4) \oplus ([x_1] \otimes y_1)]$, and is mapped to *Distribution 1* to form the asymmetry mapping $\{[([x_2 \oplus x_3] \otimes y_4) \oplus ([x_1] \otimes y_1)] \Delta\ Distribution\ 1\}$. Assuming that the distributions are for variable V, the asymmetry mapping type is a pair: $(\{T_X,T_Y\},T_V)$ where $T_k$ is the variable-type for variable $k$.

Simple braces, complex braces and asymmetry mappings are the elements in our representation. The combination operators that operate on these elements are collect and cross. We use these operators to 'factor' a CPT.

**Definition 11** Let $S_T$ be a set of elements of type T. *Collect*, $\oplus$, is a combination operator that takes two elements of the same element-type, T, to construct a complex element of the same element-type, T, such that:

(1) $S_T$ is closed under collect. For any elements x and y in $S_T$, $x \oplus y$ is in $S_T$.

(2) The atomic elements comprising $x \oplus y$ consist of the union of the atomic elements comprising x and y respectively.

(3) Collect is associative. For any elements x, y and z in $S_T$, $x \oplus (y \oplus z) = (x \oplus y) \oplus z$.

(4) Collect is commutative. For any elements x and y in $S_T$, $x \oplus y = y \oplus x$.

(5) Collect is idempotent. For any x in $S_T$, $x \oplus x = x$.

(6) The zero identity for collect is the null set. For any x in $S_T$, $x \oplus \emptyset = x$.

A complex element formed with the collect operator is called a *collection*.

A simple partition element is a collection of simple braces. Suppose that X has two possible values, $x_j$ and $x_k$, that make up a simple partition element. Then we represent that simple partition element by the complex brace $[x_j \oplus x_k]$. This brace has the same element-type as X does. On the other hand, cross produces elements with a different type. The element-types for elements being crossed must be disjoint. The element-type for a crossed pair is the set of element-types for the contributing elements. However, cross is not defined for certain pairs of elements. To properly represent conditional probability tables, a complex element's type must be a concatenation of types for the parents with the type for a single dependent variable. An asymmetry mapping already has the type for the dependent variable. Crossing two asymmetry mappings or distributions for different dependent variables is meaningless. Therefore, an asymmetry mapping may not be crossed with another asymmetry mapping and a typed distribution may not be crossed with another typed distribution..

**Definition 12** Let $S_{T1}$ be a set of elements of type T1. Let $S_{T2}$ be a set of elements of type T2. *Cross*, $\otimes$, is a combination operator that pairs two elements of disjoint element-types to produce a complex element of a third element-type such that:

(1) For any x that is a typed distribution and any y, $x \otimes y$ and $y \otimes x$ are not defined.

(2) For any x and y that are both asymmetry mappings, $x \otimes y$ and is not defined.

(3) For any elements x in $S_{T1}$ and y in $S_{T2}$ and $T1 \cap T2 \neq \emptyset$, $x \otimes y$ is not defined.

(4) For any elements x in $S_{T1}$ and y in $S_{T2}$, and $T1 \cap T2 = \emptyset$, $x \otimes y$ is in $S_{T12}$ where T12 is the concatenation of T1 and T2.

(5) Cross is associative. For any elements x in $S_{T1}$, y in $S_{T2}$ and z in $S_{T3}$, and $T1 \cap T2 \cap T3 = \emptyset$, $x \otimes (y \otimes z) = (x \otimes y) \otimes z$.

(6) Cross is commutative. For any elements x in $S_{T1}$ and y in $S_{T2}$, and $T1 \cap T2 = \emptyset$, $x \otimes y = y \otimes x$.

(7) The zero identity for cross is the null set. For any x in $S_{T1}$, $x \otimes \emptyset = x$.

(8) Cross is distributive through collect. For any x in $S_{T1}$, $y_j$ and $y_k$ in $S_{T2}$, and $T1 \cap T2 = \emptyset$, $x \otimes [y_j \oplus y_k] = [(x \otimes y_j) \oplus (x \otimes y_k)]$.

Obviously, any element in a state space for a set of parent variables can be represented by a cross of their simple partition elements. One example of $\otimes$ is a *Conditioning_Assignment*. Suppose that X and Y are parents of Z with the following simple partition elements, the complex brace $[x_j \oplus x_k]$ for X, and the simple brace $[y_i]$ for Y. The conditioning assignment is represented by the complex brace $([x_j \oplus x_k] \otimes [y_i])$. Because cross is distributive through collect, any complex brace can be reduced to a set of atomic elements from the state space. For example, $([x_j \oplus x_k] \otimes [y_i]) = [(x_j \otimes y_l) \oplus (x_k \otimes y_l)]$.

As illustrated in Figure 3, a conditioning partition element for the conditioning assignment could simply be the conditioning assignment itself or a collection of conditioning assignments.

**Definition 13** *AMap*, $\Delta$, is a map operator that pairs a brace with a distribution to construct an asymmetry mapping. The type for the distribution may not be a member of the set of types for the brace.

(1) For any elements x in $S_{T1}$ and $D_Y$ in $S_{T2}$ and $T_1 \cap T_2 \neq \emptyset$, $x \Delta D_Y$ is not defined.

(2) For any brace x in $S_{T1}$ and distribution $D_Y$ in $S_{T2}$, and $T_1 \cap T_2 = \emptyset$, $x \Delta D_Y$ is in the set $S_{T1,T2}$, all of whose elements are asymmetry mappings with type pair $\{T_1, T2\}$.

(3) Map is distributive through collect. For any elements $x_i$ and $x_j$ in $S_{T1}$, $D_Y$ in $S_{T2}$, and $T_1 \cap T_2 = \emptyset$,
$(x_i \oplus x_j) \Delta D_Y = (x_i \Delta D_Y) \oplus (x_j \Delta D_Y)$.

(4) Map is associative with cross. For any elements $x_j$ in $S_{T1}$, $y_k$ in $S_{T2}$ and $D_Z$ in $S_{T3}$, and $T_1 \cap T_2 \cap T_3 = \emptyset$,
$(x_j \otimes y_k) \Delta D_Z = x_j \otimes (y_k \Delta D_Z)$.

### 4.3 FACTORED CPTS AND COMBINATION

A factored CPT is a representation for a portion of a conditional probability table. It is a collection of asymmetry mappings. Each asymmetry mapping must represent the contribution of all of the parents for a dependent variable. Furthermore, the subset of the parent variables' state space represented by each asymmetry mapping must be disjoint from the subsets represented by the other asymmetry mappings in the factored CPT. This restricts an element of the parent state space to exactly one distribution over the values of the dependent variable. So the elements composing a factored CPT are mutually exclusive. However, they are not required to be exhaustive. This permits a factored CPT to represent the partially specified conditional probability tables of network fragments while maintaining consistency among the distributions assigned to elements of the state space for the parent variables.

**Definition 14** Let X be a random variable. Let $\Pi_X$ be the parents of X. A *factored CPT for X* is a collection of asymmetry mappings in which the braces for the elements of the asymmetry mapping are disjoint and whose element-type pair is the set of variable-types for $\Pi_X$ and the variable-type for X.

We (Mahoney, 1999) have shown that any asymmetry subnetwork and any hypothesis conditioned network fragment may be represented by a factored CPT and any factored CPT may be represented by an asymmetry subnetwork or a hypothesis conditioned network fragment. We have also shown that permitted combinations of factored CPTs produce valid factored CPTs. Furthermore, any subset of a conditional probability table can be represented by the algebra.

## 5. CONCLUSION AND FUTURE RESEARCH

We have presented an alternative representation for partially specified CPTs. The representation is first a parsimonious one for CPTs. As such, it supports the efficient elicitation and maintenance of the parameters of a Bayesian network.

It also represents the local distributions in hypothesis specific network fragments. We have also presented an algebra that supports the independence relationships and reveals context-specific independence. We described how context-specific independence can be exploited to decompose an asymmetry network into subnetworks associated with hypothesis-conditioned network fragments. We described operators for consistently combining the asymmetry subnetworks associated with hypothesis-specific network fragments during construction of a situation-specific network. This representation for partially specified CPTs makes it easy to extract variables and values that serve as context to other variables. This is not only nice for minimizing the representation of local distributions in network fragments and for making elicitation in complex domains manageable, but it is also an important tool that facilitates reasoning about context during automated network construction. This is an area that we plan to explore in a future paper. We also look towards using the algebra to extend D'Ambrosio's (1995) local expression language.

## Acknowledgements

The research reported in this paper was sponsored by DARPA and the U.S. Air Force Wright Laboratory under contract DACA76-97-0005 to Information Extraction and Transport, Inc. We also thank the anonymous reviewers for their helpful comments.